%% file: main.tex
\documentclass[10pt,twocolumn,letterpaper]{article}

\usepackage{iccv}
\usepackage{times}
\usepackage{epsfig}
\usepackage{graphicx}
\usepackage{amsmath}
\usepackage{amssymb}
\usepackage{enumitem}
\usepackage{algorithm}
\usepackage{algpseudocode}
\usepackage{booktabs}
\usepackage{multirow}
\usepackage{tikz}
\usepackage{xcolor}
\usepackage{flushend}

\newcommand*\circled[1]{\tikz[baseline=(char.base)]{
            \node[shape=circle,draw,inner sep=0.1pt] (char) {#1};}}

\usepackage[breaklinks=true,bookmarks=false]{hyperref}

\iccvfinalcopy 

\def\iccvPaperID{12033} 

\ificcvfinal\fi

\begin{document}

\title{VertexSerum: Poisoning Graph Neural Networks for Link Inference}

\author{Ruyi Ding*, Shijin Duan*, Xiaolin Xu, Yunsi Fei\\
Northeastern University,
Boston, MA, USA\\
{\tt\small \{ding.ruy, duan.s, x.xu, y.fei\}@northeastern.edu}
}

\maketitle

\def\thefootnote{*}\footnotetext{These authors contributed equally to this work. This work to be appeared in ICCV 2023.}\def\thefootnote{\arabic{footnote}}

\ificcvfinal\thispagestyle{empty}\fi

\begin{abstract}
\input{text/abstract.tex}

\end{abstract}

\input{text/introduction.tex}
\input{text/background.tex}
\input{text/motivations.tex}
\input{text/methods.tex}
\input{text/experiments.tex}

\input{text/discussions.tex}

{\small
\bibliographystyle{ieee_fullname}
\bibliography{egbib}
}

\end{document}

%% file: text/abstract.tex

Graph neural networks (GNNs) have brought superb performance to various applications utilizing graph structural data, such  as social analysis and fraud detection. The graph links, e.g.,  social relationships and transaction history, are sensitive and valuable information,  which raises privacy concerns when using GNNs. To exploit these vulnerabilities, we propose VertexSerum, a novel graph poisoning attack that increases the effectiveness of graph link stealing by amplifying the link connectivity leakage. To infer node adjacency more accurately, we propose an attention mechanism  that can be embedded into the link detection network. Our experiments demonstrate that VertexSerum significantly outperforms the SOTA link inference attack, improving the AUC scores by an average of $9.8\%$ across four real-world datasets and three different GNN structures. Furthermore, our experiments reveal the effectiveness of VertexSerum in both black-box and online learning settings, further validating its applicability in real-world scenarios.

\if false 
Graph structural data has been revolutionary in various applications, such as traffic forecasting and chemical prediction, which are often analyzed using graph neural networks (GNNs).
Unfortunately, the sensitive nature of link information, such as social relationships, raises privacy concerns when using GNNs, which are vulnerable to privacy attacks like link stealing. 
To exploit these vulnerabilities, we propose VertexSerum, a novel graph poisoning attack, that amplifies link connectivity leakage and steals graph structural privacy effectively.
We also improve the detection network with an attention mechanism to infer graph adjacency more accurately.
Our experiments demonstrate that VertexSerum significantly outperforms the baseline link inference attack, improving the AUC scores by an average of $10.1\%$ across four real-world datasets and three different GNN structures.
Furthermore, our experiments reveal the effectiveness of VertexSerum in both black-box and online learning settings, further validating its applicability in real-world scenarios.

\fi

%% file: text/introduction.tex
\section{Introduction} \label{sec: introduction}
Graph Neural Networks (GNNs) have been widely adopted in various domains, such as financial fraud detection~\cite{wang2019semi}, social network analysis~\cite{sun2021graph}, and heart-failure prediction~\cite{choi2017gram}, thanks to their capabilities to model high-dimensional features and complex structural relationships between entities \cite{zhou2020graph}. However, with the increasing use of graph data, concerns about data privacy are also growing \cite{chen2020vertically, dai2022comprehensive, wu2021fedgnn}. This is particularly relevant in industries such as finance and healthcare, where sensitive relationships are often embedded in graph-structured data.

Recently, there has been a rise in privacy attacks on GNNs~\cite{he2021stealing, wu2022linkteller} that infer the existence of links between nodes in graphs by only querying the graph model, thus posing a threat to the confidentiality of GNNs. For a graph node pair, the similarity of their posterior distributions
(abbreviated as ``posteriors'' \cite{he2021stealing}) is measured to deduce the link existence. 
For instance, in federated learning scenario~\cite{he2021fedgraphnn}, where different parties keep private data locally but contribute to the GNN training in the cloud based on their data, a malicious contributor can infer the link belonging to other contributors by querying trained GNN models. 
In this context, the risks of link information leakage lie in the joint training of GNNs and the available GNN inference APIs on graph data.


In this work, we identified a limitation of the existing link-inferring attacks: they do not perform well if the interested node pairs are from the same category (intra-class). 
This is due to the high similarity of the posterior distributions between node pairs in the same category. 
To overcome this limitation, we propose a novel approach to significantly improve link inference attacks, particularly on intra-class node pairs, by allowing a malicious contributor to poison the  graph  during GNN training in an unnoticeable way. 


This paper proposes a novel privacy-breaching data poisoning attack on GNNs, \textbf{VertexSerum}\footnote{The name is inspired by \texttt{Veritaserum} in the Harry Potter series.}, with a new analysis strategy. 
The attack aims to amplify the leakage of private link information by modifying nodes/vertices. This work makes the following contributions:
\begin{enumerate}[leftmargin=*]
    \item We propose a new evaluation metric, intra-class AUC score, for link inference attacks, by considering only node pairs from the same class. This new metric overcomes the bias of the prior works that do not differentiate between inter-class and intra-class, and brings valuable insights for our approach. 
    \item We introduce the first privacy-breaching data poisoning attack on GNNs, which injects adversarial noise into a small portion ($<$ 10\%) of the training graph to amplify the graph's link information leakage. We constructively employ a self-attention-based network to train the link detector and propose a pre-training strategy to overcome the overfitting issue of limited training data.
    \item We demonstrate the effectiveness of the proposed link inference attack on popular GNN structures and graph datasets. The attack improves the link stealing AUC score by $9.8\%$ compared to the SOTA method in~\cite{he2021stealing}.
    \item We consider the practicality of applying VertexSerum by evaluating its homophily noticeability of the poisoned graph and the victim model accuracy. The experimental results show that VertexSerum increases model privacy leakage without affecting the GNN performance.
\end{enumerate}


%% file: text/background.tex
\section{Background and Related Work} \label{sec: background}

\subsection{Graph Neural Networks} \label{sec: gnn}
Graph Neural Networks (GNNs) are widely used in semi-supervised graph node classification tasks \cite{zhou2020graph}. 
A graph, denoted as $G$=$(V, E)$, has a topology with a set of nodes $V$ and edges/links $E$. This work focuses on undirected homogeneous graphs, commonly studied in graph theory and network analysis \cite{ chen2021channel,choi2017gram, razani2021gp, sun2021graph, wang2019semi, xing2021learning}. 
A link between node $u$ and $v$ is represented by $(u,v)\in E$, while its absence is $(u,v)\notin E$.
For each node, it has features $x$ and corresponding categorical label $y$ for a classification task. Together with the graph, node features and labels compose the dataset used for GNN training and validation, denoted as $D$=$\{G, \boldsymbol{X}, \boldsymbol{Y}\}$.
After training, a neural network model for the graph is generated, denoted as $f$, where the model output $f(u)$ represents the posterior probabilities of node $u$ for the classes. 
The main GNN architectures for node classification include Graph Convolutional Network (GCN)~\cite{kipf2016semi}, Graph SAmple and aggreGatE (GraphSAGE)~\cite{hamilton2017inductive}, and Graph Attention Network(GAT)~\cite{velickovic2017graph}. These models, with different neural  network architectures, all learn to aggregate feature information from a node's local neighborhood, whose receptive field is bounded by the model depth. Different from previous works that do not differentiate between nodes in the graph for evaluation, we specifically analyze the intra-class node pairs, which refer to nodes in the same class. 


\subsection{Link Inference Attack} \label{sec: link-stealthy}
GNNs, like other machine learning models, are susceptible to various privacy attacks that compromise the confidentiality of sensitive information within the data. 
These include membership inference attacks \cite{olatunji2021membership}, adversarial graph injection attacks \cite{sun2020adversarial}, graph modification attacks \cite{Zgner2019AdversarialAO}, and link privacy attacks \cite{he2021stealing, wu2022linkteller}. 
Stealing Link Attack~\cite{he2021stealing} was the first link privacy attack, where the graph structure information is inferred from the prediction results  of the GNN model, i.e., posterior distributions of nodes. 
Another attack, LinkTeller~\cite{wu2022linkteller},  takes into account the influence propagation during GNN training for link inference. However, LinkTeller requires the attacker to have access to the graph’s node features $\boldsymbol{X}$, a much stronger attack model than ours where the attacker only accesses the posterior distributions of interested nodes, a more realistic scenario. 


\subsection{Enhance Privacy Leakage via Data Poisoning} \label{sec: enhance poisoning}
Data poisoning is an effective method to manipulate the behavior of the victim model during training by intentionally introducing malicious training samples into the benign dataset~\cite{zugner2020adversarial}. 
The recent work~\cite{chen2022amplifying} poisons the training dataset with  a small number of crafted samples, with incorrect labels, which results in a trained model that overfits the training data, significantly increasing the success rate of membership inference attacks. Inspired by the previous \textit{membership} leakage amplification by data poisoning, on conventional deep learning models,  this work shows that properly crafted data poisoning is also able to amplify \textit{link} leakage of the graph  in GNNs, posing a significant privacy threat to GNNs. 
Data poisoning on GNNs can be achieved by   modifications made to node features, node labels, or the  graph structure. 
We choose to poison node features with small perturbations to make the attack stealthy.
Our attack is more effective than the state-of-the-art link inference attacks \cite{he2021stealing, wu2022linkteller} with a specific focus on intra-class inference. 


%% file: text/motivations.tex
\section{Observations and Insights} \label{sec: motivation}

\begin{table}[t]
\resizebox{\linewidth}{!}{
\begin{tabular}{l|cc|cc} 
\toprule[1.mm]
Benchmark & $\text{R}_{linked}$ & $\text{R}_{unlinked}$ & $\text{AUC}_{all}$ & $\text{AUC}_{1}$ \\\midrule[0.6mm]
Cora & 0.81 : 0.19 & 0.18 : 0.82 & 0.907 & 0.874 \\
Citeseer & 0.74 : 0.26 & 0.18 : 0.82 & 0.987 & 0.912 \\
AMZPhoto & 0.83 : 0.27 & 0.16 : 0.84 & 0.919 & 0.813 \\
AMZComputer & 0.78 : 0.22 & 0.21 : 0.79 & 0.913 & 0.826 \\
\bottomrule[0.6mm]
\end{tabular}}
\caption{Node pairs' distribution analysis. $\text{R}$ is the ratio of intra-class node pairs vs. inter-class, among all linked node pairs and unlinked node pairs. $\text{AUC}$ reflects the success rate of link reference attacks, where $\text{AUC}_{all}$ considers overall node pairs and $\text{AUC}_{1}$ considers only node pairs from intra-class, e.g., in class 1.}
\label{tab:inter_vs_intra}
\end{table}

\subsection{Link Inference Attack Does Not Always Work} \label{sec: inter vs intra}
\begin{figure}[t]
\centering
\includegraphics[width=0.95\linewidth]{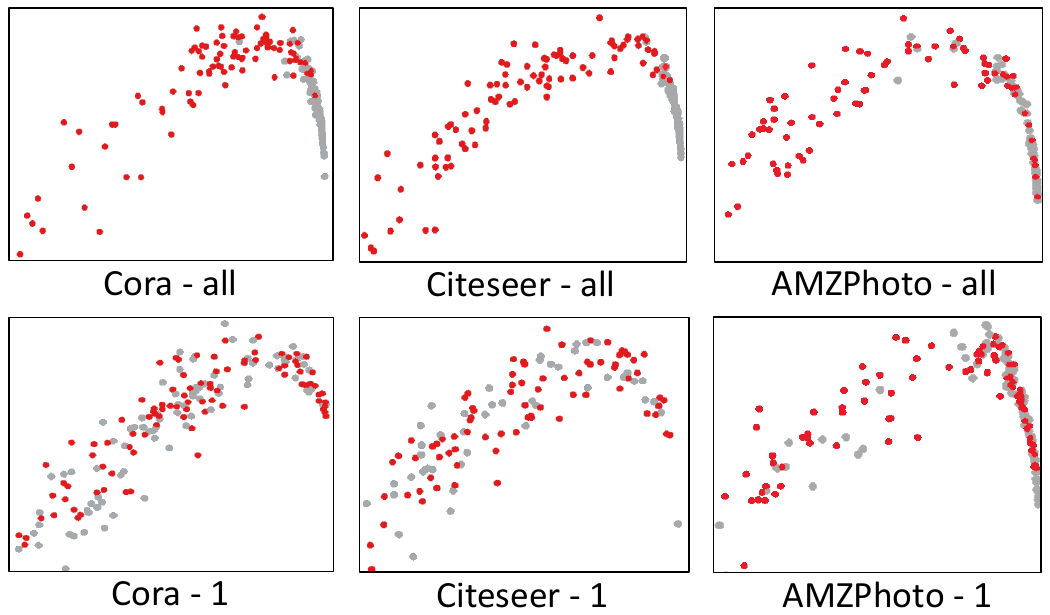}
\caption{Visualization on link inference, overall vs. intra-class. We randomly sampled 200 node pairs (100 linked + 100 unlinked) from all nodes (\textsf{all}) and only the second class (\textsf{1}). The dots are 
the PCA projections of the similarities of node pair posteriors,  where dots in \textcolor{red}{red} represent linked pairs and dots in \textcolor{gray}{gray} represent unlinked pairs. 
The more apart the two distributions are, the easier link inference can be. }
\label{fig:inter_vs_intra}
\end{figure}

Previous research of link inference attacks on GNNs has demonstrated good performance in predicting the existence of links among overall node pairs~\cite{he2021stealing}. The GNN model is queried, and the similarity of the posterior distributions of the node pair is calculated for a  link detector, which returns the prediction of whether a link exists between these two nodes. Although the performance on overall node pairs tends to be good, when considering only intra-class node pairs, i.e., to infer the link existence of node pairs from the same class, the effectiveness is much lower. This is due to several reasons: \circled{1} Though it  is common to select equal numbers of linked and unlinked node pairs for evaluation, the distribution of inter-class and intra-class node pairs in both sets are highly unbalanced: while the majority of linked node pairs are intra-class, most of the unlinked node pairs are inter-class; 
\circled{2} the posterior distributions of intra-class nodes are much more similar than those of inter-class nodes. We demonstrate the characteristic of node pairs distribution in Table \ref{tab:inter_vs_intra}. 
If we only consider node pairs from the same classes, their posterior distribution will be similar regardless of whether they are linked or not. The different success rates of the link inference attack on node pairs from the entire graph and only one class are reflected by the AUC scores, presented in the third and fourth columns of Table \ref{tab:inter_vs_intra}, and also visualized in Figure \ref{fig:inter_vs_intra}. As the visualization shows, in the top row across three different datasets, the linked node pairs and unlinked node pairs are easily distinguishable, from the \textit{overall node pairs}; while the bottom row shows that for intra-class node pairs, the two distributions are not easily separable, indicating the difficulty for link inference.  
To address this issue, we propose a new metric, \textit{intra-class AUC score}, to evaluate the link inference attack's performance in the same classes, as presented in Column 5 of Table~\ref{tab:inter_vs_intra}.

\subsection{Graph Poisoning Threat to GNNs} \label{sec: why poisoning}
Data poisoning on Graph neural networks can be achieved on various entries. For example, in social networks, an adversarial user can create fake accounts or modify their profile deliberately. 
As GNNs applied to these graphs must be frequently retrained or fine-tuned, an  attack surface is created for malicious parties to compromise the GNN performance or privacy by crafting malicious data. Specifically in federated learning, a common structural graph is used by  distributed contributors to provide data for training, malicious parties may upload carefully poisoned data into the graph in a stealthy and unobtrusive way. 
Graph poisoning attacks are easy to conduct, difficult to detect, and highly effective in compromising GNNs. Our proposed attack shows that by data poisoning, the link leakage of intra-class nodes can be significantly amplified, and link inference can be effectively accomplished.


%% file: text/methods.tex
\section{VertexSerum - The Proposed Attack} \label{sec: methods}
\begin{figure}[t]
\centering
\includegraphics[width=0.95\linewidth]{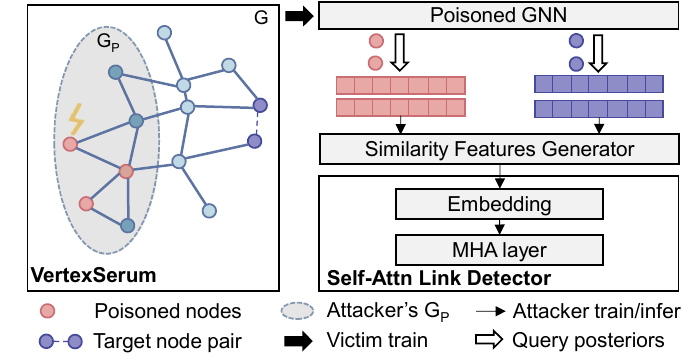}
\caption{Overview of VertexSerum with Self-Attention Detector.}
\label{fig: attack overview}
\end{figure}

In this section, we illustrate our proposed privacy-breaching data poisoning attack -- VertexSerum. 
The overview of the attack procedure is presented in Figure~\ref{fig: attack overview}.

\subsection{Threat Model} \label{sec: threat-model}

\noindent \textbf{Adversary's Goal.} 
The attack targets GNN-based classifiers, which utilize node features and the graph topology to predict the labels for querying nodes. The attacker aims to deduce the connectivity between any pair of nodes $(u, v)$ belonging to class $k$ by querying a pre-trained GNN model. 

\noindent  \textbf{Adversary's Knowledge.}
We assume the attacker has limited access to the vendor's GNN as they can only acquire the interested nodes' posteriors through queries. We also assume that the attacker has access to a portion of the graph, as in federated learning, where the attacker acts as a distributed contributor to provide data for the training dataset, which can be intentionally poisoned. Note these assumptions align with ``Attack-3'' in the state-of-the-art link inference attack~ \cite{he2021stealing} and are practical. 
We limit the portion of the graph that the attacker can manipulate, such as $10\%$ of the entire graph, which is more practical and realistic in federated learning settings.


\subsection{Inspiration from ML Poisoning} \label{sec: intuition}
In conventional machine learning (ML) regime, poisoning the training dataset with tainted data can expose user data privacy \cite{chen2022amplifying, tramer2022truth}, e.g., by injecting label-tampered samples to the training data, forcing the victim model to overfit on specific features of each sample, thereby exacerbating its membership leakage. 
However, the potential of such data poisoning schemes have not been explored in attacking the link privacy of GNNs. This work bridges this knowledge gap by 
crafting samples in training dataset to strengthen GNN model's attention on node connections, making the model to produce more similar outputs for linked nodes and increase the dissimilarity between unlinked nodes. Rather than generating abnormal labels which may be detected by outlier detection tools, we induce poisoned features with small perturbations with Projected Gradient Descent (PGD), allowing us to achieve attack stealthiness.

\subsection{Attack Flow of VertexSerum} \label{sec: edge privacy game}

\begin{algorithm}[t]
\caption{Link Stealing with VertexSerum}\label{alg: VertexSerum}
\begin{algorithmic}[1]

\Require Target class $k$; 
Partial graph $G_p=(V_p, E_p)$;
Step size $\epsilon$; 
Maximum number of iterations $N$;
Training algorithm $\mathcal{T}$;
Vendor graph $G=(V,E)$.
\Ensure Link existence of node pair $\hat{z}=(u,v)$.
\Statex \textcolor{gray}{/*Generate Poisoned Graph*/}
\State Train shadow GNN model on $G_p$ with the public training algorithm $f^{\text{sh}}_{\theta}\leftarrow \mathcal{T}(G_p)$. \label{line: start poisoning}
\For{$n =1,2, \dots, N$}\Comment{\textcolor{teal}{Projected Gradient Descent}}
\State Compute the gradient of loss $L$($f^{\text{sh}}_{\theta}$): $g_n \gets \nabla L$
\State Update nodes features in class $k$ to increase loss, $x \in V^k_p$: $x_{n+1} \gets x_n + \epsilon g_n$
\EndFor
\State Get poisoned graph $G'_p$ and send it to the vendor. \label{line: end poisoning}
\State The vendor trains a GNN model $f_{\theta}$ on $G\cup G'_p$.

\Statex \textcolor{gray}{/*Train Link Detector*/}
\State Query model $f_{\theta}$ to obtain posteriors of nodes in $V^k_p$. \label{line: start train ld}
\State Compute and aggregate the similarity features $\boldsymbol{F}^k_p$ from node pairs' posteriors as truth dataset $\mathcal{D}^k_p=\{\boldsymbol{F}^k_p, E^k_p\}$.
\State Train binary classifier $\mathcal{M}$ on $\mathcal{D}^k_p$ using self-attention link detector.  \label{line: end train ld}

\Statex \textcolor{gray}{/*Link Inference*/}
\State Given a target node pair in class $k$, $u, v \in V^k$, compute the similarity feature $\boldsymbol{F}^k_{u,v}$. \label{line: start attack}
\State Feed in $\boldsymbol{F}^k_{u,v}$ to detector $\mathcal{M}$ for link inference.
\State \textbf{return True / False} \label{line: end attack}
\end{algorithmic}
\end{algorithm}

VertexSerum aims to steal the true link information of interested node pairs. The attack is carried out between a model vendor $\mathcal{V}$ and a malicious contributor $\mathcal{A}$. 
The vendor has access to the entire graph dataset $D$=$\{G, \boldsymbol{X}, \boldsymbol{Y}\}$ and trains a downstream task with a public training algorithm $\mathcal{T}$ \footnote{We assume the GNN type is open to the adversary for the ease of evaluation. We also demonstrate the effectiveness of VertexSerum in Section \ref{sec: blackbox}, when the adversary has no clue of the GNN model.}. 
The adversary contributes a small portion of the dataset, $D_p$=$\{G_p, \boldsymbol{X}_p, \boldsymbol{Y}_p\}$, containing a partial graph $G_p$, which is used for both generating the poisoning sub-graph and training the link detector.
The attack steps are:
\begin{enumerate}[leftmargin=*]
    \item The adversary chooses a target class $k$ from the label space $\boldsymbol{Y}$.  The attack goal is to predict the link existence between nodes $u, v$, i.e., if $(u, v) \in E$, when $y_u$=$y_v$=$k$.
    
    \item Following the steps in Lines~\ref{line: start poisoning}-\ref{line: end poisoning} of Algorithm~\ref{alg: VertexSerum}, the adversary \textbf{\textit{generates a partial dataset $D'_p$ with a poisoned graph $G'_p$}} by analyzing a shadow model trained on $G_p$, as depicted in the shadow part in Figure~\ref{fig: attack overview}, and \textbf{\textit{sends it to the vendor}}.
    
    \item The vendor trains a GNN model for downstream tasks $f_{\theta}\leftarrow \mathcal{T}(D \cup D'_p)$ on the poisoned graph $G\cup G'_p$.

    \item The adversary queries the GNN  model, $f_{\theta}$, with the possessed poisoned partial graph $G'_p$ and generates similarities of posteriors. \textbf{\textit{Binary link detectors}} are constructed to infer link existence, as shown in the right bottom part of Figure~\ref{fig: attack overview} and detailed in Lines~\ref{line: start train ld}-\ref{line: end train ld} of Algorithm~\ref{alg: VertexSerum}.
    \item The adversary makes a guess $\hat{z} ={(u, v)}$ with the link detectors (Line~\ref{line: start attack}-\ref{line: end attack}).
\end{enumerate}
Our attack utilizes data poisoning to breach the confidentiality of GNNs: the poisoned graph $G'_p$ is used in the victim GNN model training, with an objective to amplify the model privacy leakage. 

\subsection{Requirements of the Poisoning Nodes} \label{sec: requirments}
For Step 2 of the attack, to generate a graph that enhances the model's aggregation on linked nodes, we design a specific poisoned graph $G'_p$ that makes the GNN model $f_{\theta}$ focus more on adjacency.
Next, we outline requirements for successful node poisoning:
\begin{enumerate}[leftmargin=*]
    \item \textbf{Intact Community.}
    The adversary should ensure that the node classification accuracy for the victim task is not evidently affected, so that the poisoned graph is less likely to be rejected by the vendor for GNN training. Besides, misclassified nodes can negatively impact passing information to adjacent linked nodes, leading to an overall lower aggregation capability for the GNN model.
    
    \item \textbf{Node Attraction and Repulsion.} 
    The poisoned samples should simultaneously promote the similarity of the GNN outputs on linked nodes (attraction) and the dissimilarity on unlinked nodes (repulsion). This requires a balance between the attraction and repulsion of node features when poisoning the dataset.
    
    \item \textbf{Adversarial Robustness.}
   Adversarial training techniques~\cite{shafahi2019adversarial, tramer2017ensemble} can improve a model's robustness against adversarial samples, where the model tolerates small input perturbations and outputs similar predictions. 
    In VertexSerum, we utilize adversarial training to increase the model's adversarial robustness, guiding linked nodes with similar features to produce similar posteriors.
\end{enumerate}

\subsection{Crafting Poisoning Features via PGD} \label{sec: poisoning method}
To meet these requirements, we propose a graph poisoning method optimized with projected gradient descent (PGD). 
We adopt the shadow training methods~\cite{he2021stealing, shokri2017membership}, where the attacker will first train a shadow GNN ($f_\theta^{sh}$) on the possessed partial graph $G_p$. 
The optimal perturbation to add on node features is found based on the gradient of the loss function shown in Eq. \ref{eq: loss function}.
\begin{equation} \label{eq: loss function}
\small
    L =  \alpha L_{attraction} + \beta L_{repulsion} + \lambda L_{CE} 
\end{equation}

The loss function includes three terms, with $\alpha, \beta, \lambda$ as positive coefficients to balance  attraction and repulsion:
\begin{enumerate}[leftmargin=*]
    \item The attraction loss penalizes the euclidean distance of posteriors on two linked nodes.     The PGD will find node features that reduce the distance between linked nodes.
    \begin{equation}\label{eq: attraction loss}
    \small
        L_{attraction} = -\sum_{(u,v) \in E} (f_\theta^{sh}(u) - f_\theta^{sh}(v))^2
    \end{equation}

    \item The repulsion term computes the cosine similarity between unlinked nodes. The rationale is that cosine is bounded so as to avoid an overlarge dissimilarity term. The PGD will find the node features that reduce the similarity between unlinked nodes.
    \begin{equation} \label{eq: repulsion}
    \small
        L_{repulsion} = \sum_{\substack{u, v \in V, u \not= v, \\ (u, v) \not\in E}} (1 - cos(f_\theta^{sh}(u), f_\theta^{sh}(v)))^2 
    \end{equation}

    \item The cross-entropy term $L_{CE}$ serves as a regularization in the loss function. 
    Its goal is to improve the victim model's adversarial robustness to amplify link leakage.
\end{enumerate}
The previous poisoning attack includes regularization of perturbations, such as the L1 norm, during optimization. However, we observed that this term is not necessary for the PGD process if we have a small updating step size $\epsilon$. By only optimizing Eq. \ref{eq: loss function}, the generated perturbation is already effective and unnoticeable.

\subsection{Self-attention Link Detector} \label{sec: link detector}



In Step 4 of the attack, the adversary trains a link detector using the posteriors of the partial graph by querying the pre-trained vendor model. 
Previous work~\cite{he2021stealing} used a Multi-Layer Perceptron (MLP) to analyze the similarity features of the node pair posteriors. 
However, the dense structure of MLP is often inadequate to capture the complex dependencies among similarity features. 
Furthermore, since the attacker only has a small part ($<10\%$) of the graph, training an MLP is prone to be unstable due to overfitting. 
Moreover, since VertexSerum introduces more complex characteristics such as attraction and repulsion during poisoning, the underlying patterns in the similarity features are expected to be more informative.
To address these issues, we propose  improvement to the MLP model with a Multihead Self-attention~\cite{vaswani2017attention} link detector, which can efficiently use information by selectively attending to different parts in the input similarity features. 
We follow the same construction of similarity features as the previous method~\cite{he2021stealing}, consisting of eight distances and four entropy features between two nodes. 
To ensure stability of the self-attention detector on a small dataset, we initialize its first embedding layer with the first fully-connected layer from the MLP. 
The experimental results in Table \ref{tab: attack performance} in next section show that the introduction of self-attention improves the attack AUC score by an average of $7.2\%$ with the standard deviation dropping by $35\%$. 

%% file: text/experiments.tex
\section{Experiments} \label{sec: experiments}

\subsection{Experimental Setup} \label{sec: experiment setup}

\noindent \textbf{Datasets:}
We evaluate the effectiveness of VertexSerum on four publicly available datasets: Citeseer~\cite{kipf2016semi}, Cora~\cite{kipf2016semi}, Amazon Photo Dataset~\cite{mcauley2015image}, and Amazon Computer Dataset~\cite{mcauley2015image}.
These datasets cover different daily-life scenarios and are widely used as benchmarks for evaluating graph neural networks.
The first two datasets are citation networks where nodes represent publications, and links indicate citations among them. The last two datasets are co-purchase graphs from Amazon, where nodes represent products, and edges represent the co-purchased relations of products. 
Our benchmarks scale from (3k nodes + 11k edges) for Cora to (14k nodes + 492k edges) for AMZComputer.
We assume the vendor's model is trained on $80\%$ of the nodes and evaluated on the remaining in the graph. 

Since we assume the attacker only contributes a small portion of the graph for training, i.e., $G'_p$, we sample $10\%$ nodes among the training dataset.
To train the link detector, we collect all linked node pairs and randomly sample the same number of unlinked node pairs in $G'_p$. Similarity features are computed based on these node pairs, following \cite{he2021stealing}, together with corresponding link information. 
We split this dataset into $80\%$ for training and $20\%$ for validation.

\noindent \textbf{Metric:}
ROC-AUC is a commonly used evaluation metric for binary classification tasks and has also been applied in previous works on link inference~\cite{he2021stealing, wu2022linkteller}. 
It measures the ability of the link detector to distinguish between linked and unlinked node pairs. 
A higher AUC indicates superior performance of the link detector in identifying linked node pairs from unlinked ones.

In addition to overall AUC, we also evaluate the intra-class AUC. 
Overall AUC measures the ability of the link detector to identify linked node pairs among all classes, while intra-class AUC measures its ability only in one class. 
As mentioned in Section \ref{sec: inter vs intra}, a successful link inference attack should have a high overall AUC as well as a high intra-class AUC. Without loss of generality, we set Class 1 as target class to evaluate performance of the link inference attack.

\noindent \textbf{Models:}
We evaluate VertexSerum on three commonly used GNN structures: GCN~\cite{kipf2016semi}, GraphSAGE~\cite{hamilton2017inductive}, and GAT~\cite{velickovic2017graph}.
Deep Graph Library (DGL) is used for model implementation~\cite{wang2019deep}. 
We construct a 3-layer MLP as the baseline link detector, with the first layer containing 64 hidden neurons which is also the initialization for the self-attention link detector. 
The self-attention detector is of a 16-head attention structure with an input dimension of 64. 
For initialization, we train MLP for 50 epochs with a learning rate of 0.001.
We then fine-tune the self-attention detector with a learning rate of 0.0001,
using the cross-entropy loss and Adam optimizer \cite{kingma2014adam}.
We run experiments 10 times and report the average and standard deviation of AUC scores.

\begin{table*}[t]
    \centering
    \begin{tabular}{l|l|l|l|l|l|l}
    \toprule[1mm]
     Model & \multicolumn{2}{c|}{GCN} & \multicolumn{2}{c|}{GAT} & \multicolumn{2}{c}{GraphSAGE} \\
    \midrule[0.6mm]       
     Dataset & Citeseer & Cora & Citeseer& Cora & Citeseer& Cora \\
     \hline
     \multicolumn{1}{l|}{SLA + MLP\cite{he2021stealing}}    & 0.914$\pm$0.008 & 0.874$\pm$0.018 &  
        0.969$\pm$0.002 & 0.845$\pm$0.011 &  
        0.972$\pm$0.002 & 0.854$\pm$0.009 \\
    \hline
    \multicolumn{1}{l|}{SLA + ATTN }     & 0.951$\pm$0.064 & 0.903$\pm$0.067 & 
        0.980$\pm$0.003 & 0.868$\pm$0.029 & 
        0.976$\pm$0.007 & 0.931$\pm$0.029\\
    \hline
     \multicolumn{1}{l|}{VS + MLP }     & 0.892$\pm$0.006 & 0.912$\pm$0.065 &  
        0.913$\pm$0.005 & 0.856$\pm$0.017 &
        0.949$\pm$0.007 & 0.859$\pm$0.027\\
    \hline
     \multicolumn{1}{l|}{VS + ATTN(*)}     & \textbf{0.978$\pm$0.033} & \textbf{0.927$\pm$0.023} &  
        \textbf{0.997$\pm$0.002} & \textbf{0.924$\pm$0.022} &
        \textbf{0.994$\pm$0.006} & \textbf{0.957$\pm$0.007} \\
    \midrule[0.6mm]              
     Dataset & AMZPhoto & AMZComputer & AMZPhoto& AMZComputer & AMZPhoto& AMZComputer \\
     \hline
    \multicolumn{1}{l|}{SLA + MLP\cite{he2021stealing}}    & 0.813$\pm$0.015 & 0.826$\pm$0.018&  
        0.881$\pm$0.007 & 0.820$\pm$0.046 &  
        0.873$\pm$0.015 & 0.883$\pm$0.004 \\
    \hline
    \multicolumn{1}{l|}{SLA + ATTN }     & 0.917$\pm$0.037 & 0.956$\pm$0.007 & 
       0.963$\pm$0.011 &  0.889$\pm$0.066& 
       0.972$\pm$0.009& 0.978$\pm$0.005\\
    \hline
     \multicolumn{1}{l|}{VS + MLP }     & 0.780$\pm$0.007 & 0.849$\pm$0.009 &  
       0.917$\pm$0.006 &  0.852$\pm$0.033&
        0.873$\pm$0.032 & 0.898$\pm$0.004\\
    \hline
     \multicolumn{1}{l|}{VS + ATTN(*)}     & \textbf{0.939$\pm$0.018} & \textbf{0.962$\pm$0.011} &  
       \textbf{0.990$\pm$0.008} & \textbf{0.919$\pm$0.031} &
       \textbf{0.987$\pm$0.006} & \textbf{0.985$\pm$0.006}\\

    \bottomrule[0.6mm]
    \end{tabular}
    \caption{Comparison of the average AUC with standard deviation for different attacks on the four datasets. 
    The best results are highlighted in bold. (*) denotes our proposed method.
    }
    \label{tab: attack performance}
\end{table*}

\subsection{Graph Visualization} \label{sec: graph visualization}
\begin{figure}[t]
  \centering
  \includegraphics[width=0.94\linewidth]{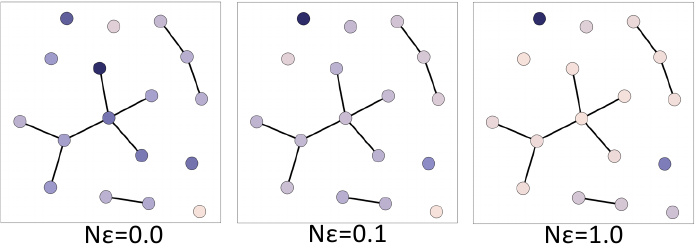}
  \caption{A visualization of nodes and edges belonging to the target class from the original ($N\epsilon=0$) and poisoned ($N\epsilon>0$) partial graphs.
  Node color represents the low-dimensional embedding of the GNN model's output, i.e., the node posteriors. 
 \textbf{Color's similarity} indicates \textbf{posteriors' similarity}.}
  \label{fig: node embedding}
\end{figure}

Figure \ref{fig: node embedding} displays part of the poisoned graph of VertexSerum on a 3-layer GraphSAGE model trained on the Cora dataset with different distortions $N\epsilon$. 
By injecting poisoned samples into the partial graph while maintaining the topology, the PGD objective loss induces corresponding attraction and repulsion forces between nodes,
resulting in increased attention to linked nodes.
As the distortion increases from $0$ to $1$, the node colors shift to demonstrate attraction to linked nodes and repulsion to unlinked nodes. 

\subsection{Attack Performance} \label{sec: attack performance}
We evaluate the effectiveness of VertexSerum (VS), including both the poisoning method and the self-attention-based (ATTN) link detector. The prior stealing link attack (SLA)~\cite{he2021stealing} serves as the SOTA method for us to compare, as it shares a similar threat model with our attack.
SLA uses similarity features and an MLP-based link detector to attack a graph neural network, without poisoning.
We compare the performance of different attack strategies and link detector structures,
and report intra-class AUC scores in Table~\ref{tab: attack performance}.

VertexSerum with the attention detector significantly improves the performance of link inference attacks for all datasets and GNN models. Compared to the method using SLA with MLP, our attack has an average improvement of $9.8\%$ on AUC scores. Note that the self-attention-based link detector significantly improves the attack performance even without poisoning datasets (see the two rows  of ``SLA + ATTN '' in Table \ref{tab: attack performance}). This is because the multi-head attention structure models the dependencies between elements in similarity features, better exposing the link existence during inference. On the other hand, using VertexSerum with MLP alone does not improve the detection performance on some datasets, such as Citeseer and AMZPhoto. From our consideration, VertexSerum enforces GNN to learn more about the connections between nodes, adding more hidden information to the similarity feature, for which MLPs lack the capability to capture. However, by combining VertexSerum with our proposed self-attention link detector, the poisoning works effectively towards increasing the link leakage.

\begin{figure}[t]
\centering
\includegraphics[width=0.95\linewidth]{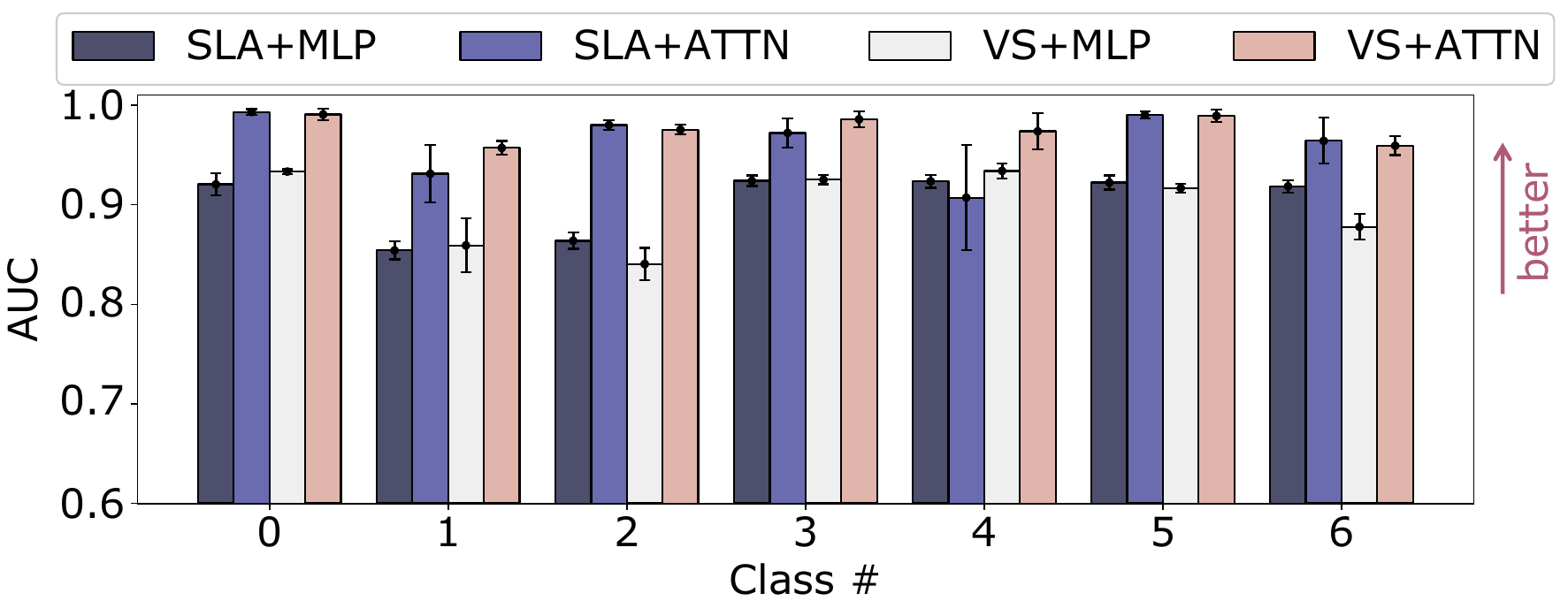}
\caption{The AUC score along each target class. We take a case study on Cora dataset (7 classes in total) with GraphSAGE as the GNN model.}
\label{fig:overall}
\end{figure}

We also demonstrate the intra-class AUC scores by varying the target class, taking the Cora dataset with the GraphSAGE model in Figure \ref{fig:overall} as an example. We can draw the same conclusion as above on the link inference attack. Not only the self-attention detector can greatly outperform the MLP detector, but the poisoning also boosts link detection as well.
Further, we demonstrate that VertexSerum can still preserve the highest effectiveness of link inference on overall classes. We show the overall AUC scores in Table \ref{tab:overall_comparison}, assuming the GNN model is based on GraphSAGE. Besides the elevated attack success, we can explicitly observe the overall AUC scores are higher than the intra-class AUC scores. This also affirms our observation discussed in Section \ref{sec: inter vs intra} that evaluation on overall node pairs yields higher performance than that on intra-class node pairs.


\begin{table}[t]
\resizebox{\linewidth}{!}{
\begin{tabular}{l|cccc}
\toprule[1mm]
         & Cora & Citeseer & AMZPhoto & AMZComputer \\\midrule[0.6mm]
SLA+MLP \cite{he2021stealing}  &  0.907$\pm$0.001&  0.987$\pm$0.001& 0.919$\pm$0.020& 0.913$\pm$0.043 \\
SLA+ATTN & 0.994$\pm$0.008 & \textbf{0.995$\pm$0.001} & 0.947$\pm$0.005 & 0.962$\pm$0.005 \\
VS+MLP   & 0.945$\pm$0.003 & 0.978$\pm$0.013 & 0.946$\pm$0.010 & 0.900$\pm$0.055 \\
VS+ATTN  & \textbf{0.997$\pm$0.012} & 0.994$\pm$0.001 & \textbf{0.956$\pm$0.001} & \textbf{0.968$\pm$0.004} \\\bottomrule[0.6mm]
\end{tabular}}
\caption{Comparison of the overall AUC scores for different tasks on GraphSAGE model, by inferring the link between node pairs from all classes.}
\label{tab:overall_comparison}
\end{table}

\begin{figure}[t]
\centering
\includegraphics[width=\linewidth]{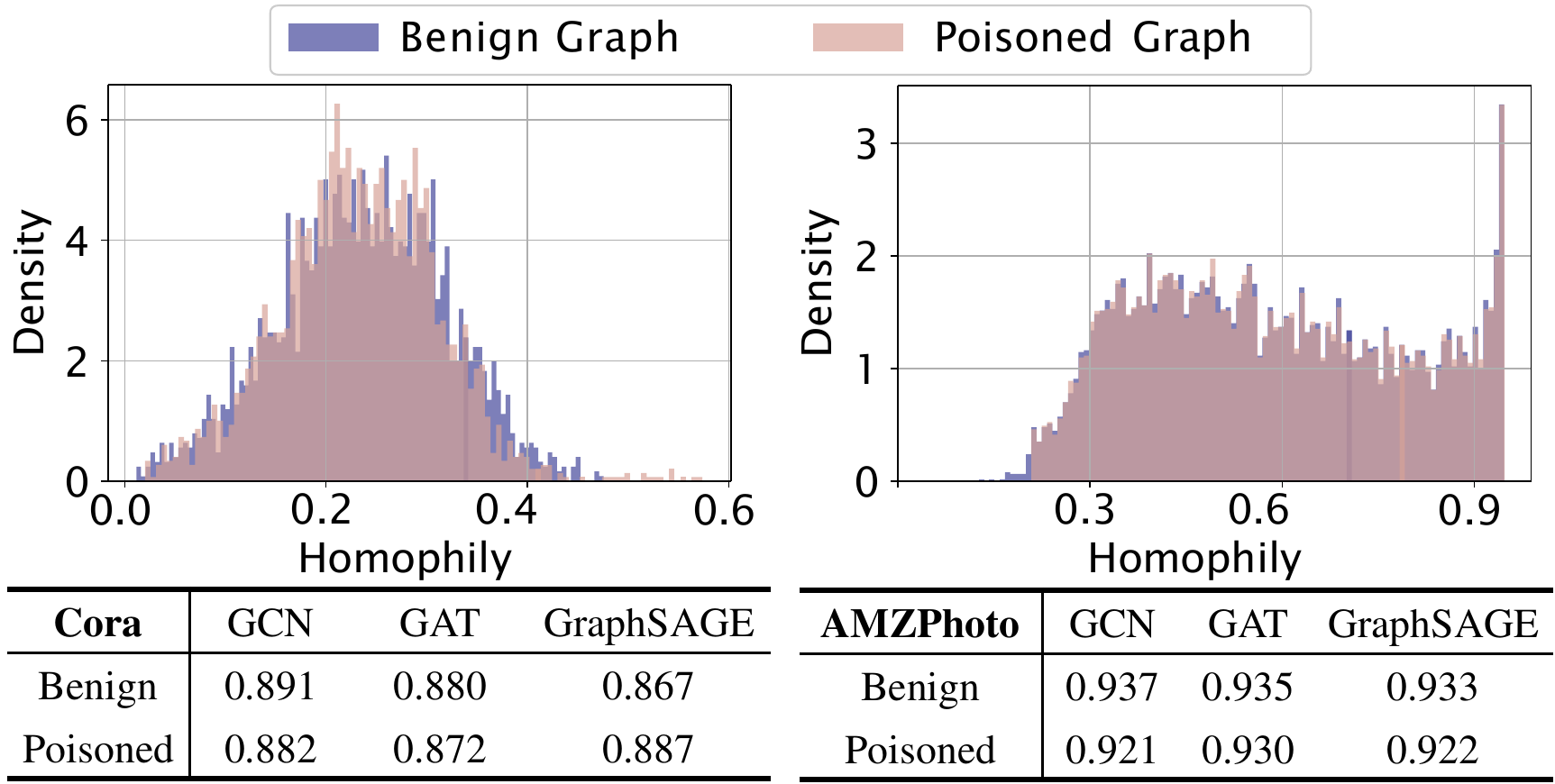}
\caption{Homophily analysis on graph poisoning. Cora and AMZPhoto are selected as the case study. The top histogram plots show the node homophily before and after the poisoning attack, where high coincidence on distribution means two graphs have high homophily. The lower tables demonstrate various model accuracies on the graphs before and after poisoning, showing that the accuracy is barely affected by the poisoning.}
\label{fig:homophily}
\end{figure}

\subsection{Attack Stealthiness} \label{sec: homophily unnoticeability}
We evaluate the stealthiness of VertexSerum from two perspectives: homophily unnoticeability and model accuracy. Homophily unnoticeability is an important metric for graph adversarial attacks and is defined as the node-centric homophily distribution shifting between the clean and poisoned graph being upper-bounded by a threshold, which ensures that the malicious nodes are not easily detectable by the database administrators~\cite{chen2022understanding}.
We visualize the homophily distribution of the benign and poisoned graphs in Figure~\ref{fig:homophily}. 
It is clear that VertexSerum can effectively preserve the homophily while still conducting effective poisoning. 
The lower tables in Figure~\ref{fig:homophily} present the model accuracy before and after poisoning, demonstrating that VertexSerum only introduces small accuracy degradation/improvement. Since from the vendor's perspective, the new accuracy is achieved after the re-training, thus, the trivial difference ensures stealthiness, i.e., the vendor will not stop using the poisoned graph due to poor performance.

\subsection{Ablation Study} \label{sec: ablation study}

\subsubsection{Influence of the Depth of GNN} \label{sec: over-smoothness}
\begin{figure}[t]
\centering
\includegraphics[width=\linewidth]{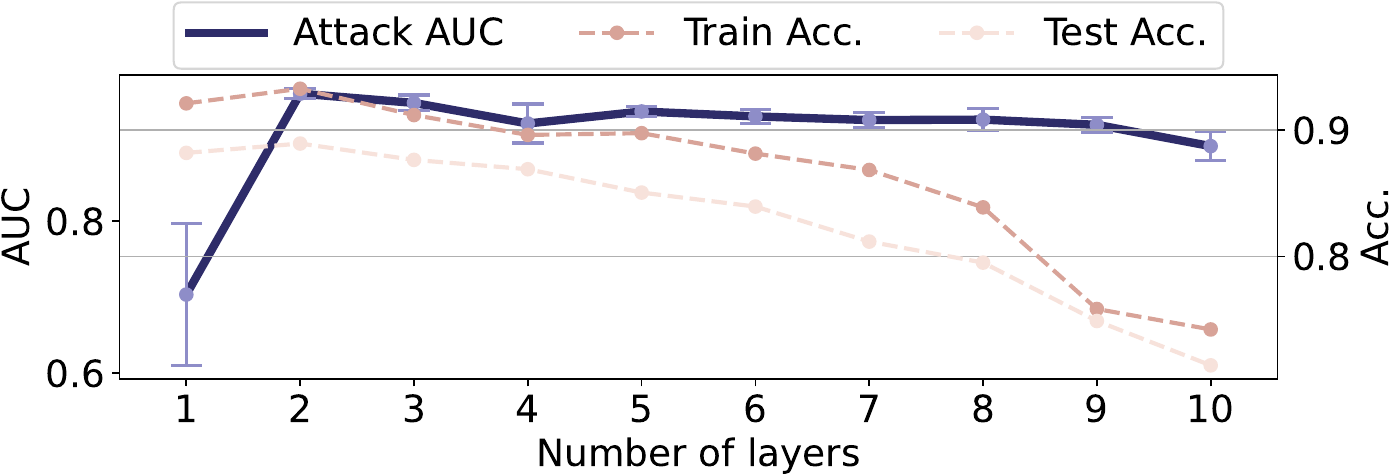}
\caption{Performance of our attack on the GraphSAGE model with varying numbers of layers.
The blue line represents the attack AUC scores, while the pink dashed lines indicate the training and testing accuracies.}
\label{fig: oversmooth}
\end{figure}

We conduct an evaluation of our attack  on the GraphSAGE model with varying numbers of layers (depth) in the GNN $f_{\theta}$.
The results are shown in Figure~\ref{fig: oversmooth}, where the blue line illustrates the attack AUC scores, while the pink dashed lines indicate the training and testing accuracy.
As the number of layers increases, the GNN aggregates information from neighborhoods across multiple hops progressively, leading to overly similar output representations on linked nodes, known as over-smoothing \cite{chen2020measuring}.

When GNNs have only one layer, the attack is harder because of the lack of aggregated information between linked nodes. VertexSerum shows good performance when the number of layers is greater than 1, as more hops of neighbors are taken into consideration. Meanwhile, the model training and testing accuracy decreases as the number of layers increases, because of over-smoothing, where the representations of nodes become similar after multi-layer message passing. Consequently, the attack performance slightly drops, due to the underperformance of model accuracy. This is a concerning observation since the attack success rate is bound to the model accuracy. A well-performed model is also highly vulnerable to link inference attacks.


\subsubsection{Impact of Different Loss Terms} \label{sec: impact of loss}
\begin{table}[t]
    \centering
    \resizebox{0.9\linewidth}{!}{
    \begin{tabular}{l|cc|cc|cc}
    \toprule[1mm]
         &  \multicolumn{2}{c|}{$\beta$ = $0.01$} & \multicolumn{2}{c|}{$\beta$ = $0.1$} & \multicolumn{2}{c}{$\beta$ = $1$} \\
    \cline{2-7}
          & $\lambda$ = $0.1$ & $\lambda$ = $1$  & $\lambda$ = $0.1$ & $\lambda$ = $1$ & $\lambda$ = $0.1$  & $\lambda$ = $1$ \\
    \hline
        $\alpha$ = $0.1$  & 0.914 & 0.942 & 0.931 & 0.943 & 0.954 & 0.953 \\
    \hline
        $\alpha$ = $1$ & 0.952 & \textbf{0.963} &  0.954 & 0.953 & 0.946 & 0.945 \\
    \hline 
        $\alpha$ = $10$ & 0.949 & 0.947 & 0.950 & 0.949 & 0.925 & 0.926 \\
    \bottomrule[0.6mm]
    \end{tabular}}
    \caption{AUC scores of VertexSerum Attack on GraphSAGE for Cora Dataset with different regularization strengths.}
    \label{tab: regularization}
\end{table}

In designing our PGD objective loss in Eq. \ref{eq: loss function}, we consider a trade-off between the attraction loss, repulsion loss, and cross-entropy loss by controlling the corresponding regularization strength terms $\alpha, \beta,$ and $\lambda$.  
We compare the attack performance using different tuples of regularization weights in Table~\ref{tab: regularization}. 
We find that the optimal choice is $(\alpha, \beta, \lambda)$=$(1, 0.01, 1)$, where the repulsion weight is much smaller than the others. 
This is due to the imbalance between the number of linked and unlinked node pairs, which leads to a high repulsion loss, and this choice balances the effect of the repulsion loss and attraction loss.

\subsection{Online Poisoning on GNNs} \label{sec: online}
\begin{figure}[t]
\centering
\includegraphics[width=\linewidth]{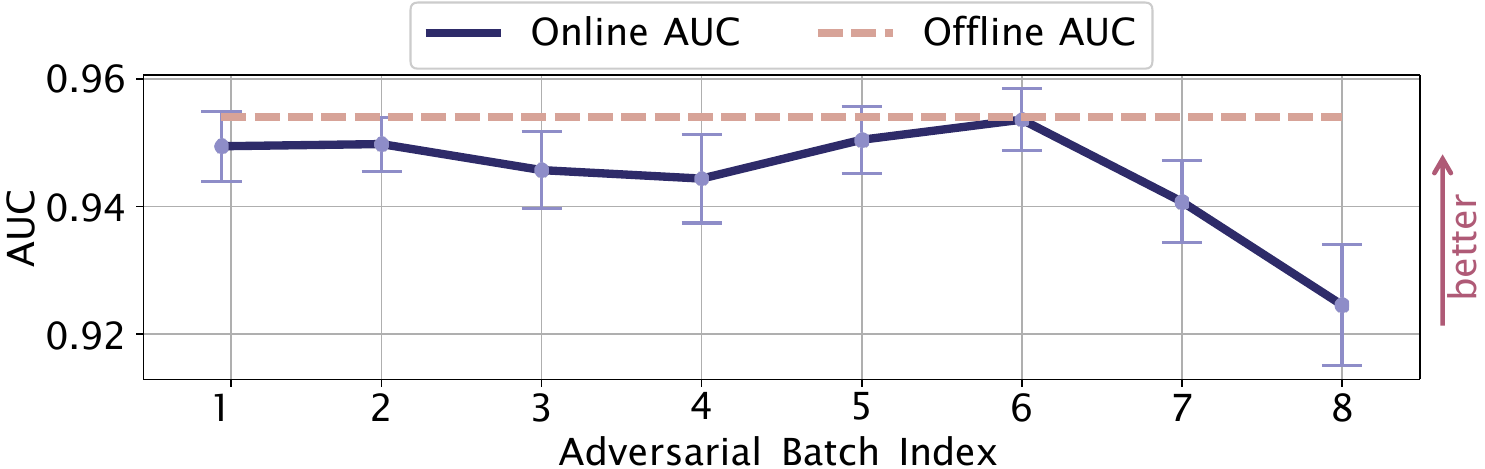}
\caption{Performance of our attack on the GraphSAGE model under the online training setting.
The blue line in the plot represents the attack AUC scores, and the x-axis represents different poisoning time during online training.}
\label{fig: online}
\end{figure}

Graph neural networks in practice are not always trained offline, but multiple contributors may provide data at different times for online training.  This is particularly relevant in scenarios such as recommendation systems, where models are frequently updated with incoming user behavior data.
In this section, we investigate a 
training scenario where the vendor's model is trained batch-by-batch as the data arrives. 
We divide the dataset into eight batches, each representing a different contributor.
We select one of the contributors as the adversary and use VertexSerum to poison the corresponding partial graph.
The model is updated in order as the contributors arrive, and we evaluate the attack performance when the adversarial contributor arrives at different times.

Figure~\ref{fig: online} presents the attack AUC when the adversarial batch arrives at different times during online training. 
We observe that poisoning the early batches is more effective than poisoning the last batch. 
This is likely because the early batches have a long-term effect on fitting the online model, while the poisoning data in the last round is only fitted during the last update. 
Further, the poisoning attack on offline training yields better results.
Since the poisoning exists throughout the offline training, the model fitting on the benign batches is also consistent throughout the training, akin to poisoning at an early time.
 Overall, VertexSerum is effective for both online and offline training on GNNs.

\subsection{Transferability in the Black-Box Setting} \label{sec: blackbox}

\begin{figure}[t]
\centering
\includegraphics[width=0.9\linewidth]{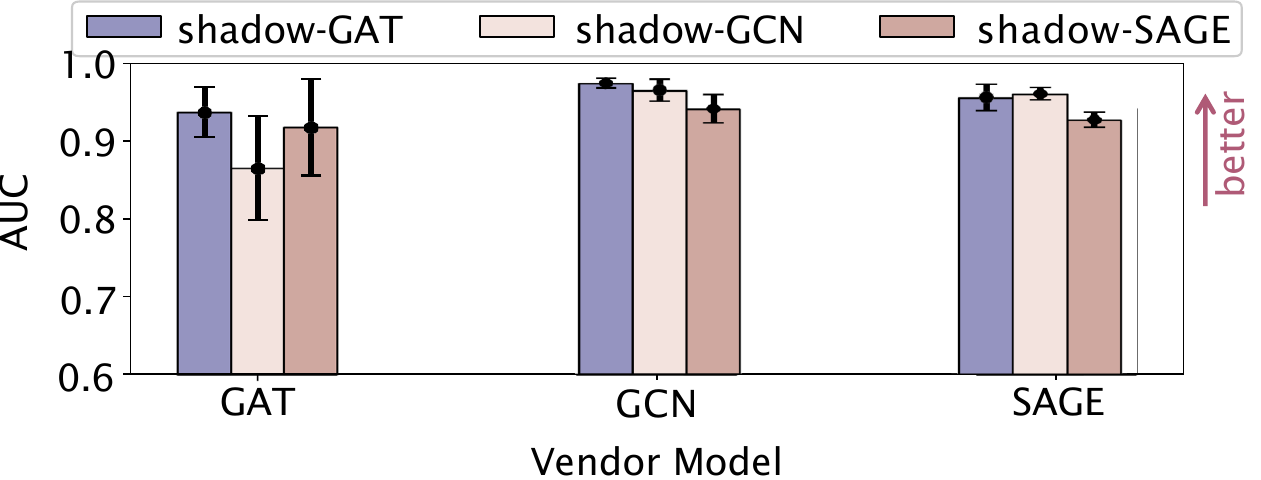}
\caption{The attack performance when the vendor model is unknown and trained on Cora Dataset, where the attacker uses arbitrary GNN structures to train the shadow model.}
\label{fig: blackbox}
\end{figure}

In previous evaluations, we assume that the attacker has prior knowledge of the vendor model's architecture and training process, which is a gray-box setting.
In this section, we extend our evaluation to the black-box setting, where the attacker has no knowledge of the victim model's architecture and configuration. 
We investigate the transferability of VertexSerum, where the attacker trains the subgraph using a different model from the vendor model.
For instance, the attacker may train the subgraph using GAT when the vendor model is trained using GraphSAGE.
Figure~\ref{fig: blackbox} shows the results under the black-box setting.
We find that even without knowledge of the vendor model structure, the attacker can still achieve high performance using VertexSerum.
Interestingly, the attacker achieves the highest AUC scores when using GAT as the shadow model to generate the poison example. 
We hypothesize that GAT has a higher generalizability in estimating the real boundary of the vendor model, making the poison samples from GAT more effective.

%% file: text/discussions.tex
\section{Defense} \label{sec: discussion}
There are two potential directions to defend against the VertexSerum attack.
The first approach is to blur the perturbation. 
Our poisoning samples are similar to adversarial samples, which are clean features with small added noise. 
Thus, it is possible to slightly change the training samples through preprocessing methods such as denoising or augmentation, without harming the model accuracy.
The second approach is to increase the GNN's robustness against the link stealthy attack. 
One way to achieve this is to build GNNs with certified robustness using differential privacy \cite{gao2020protecting}. 
Alternatively, the vendor can train the GNN with an appropriate depth to avoid over-smoothing or over-fitting.

\section{Conclusions} \label{sec: conclusion}
In this paper, we investigate the vulnerability of graph neural networks to privacy leakage amplified by data poisoning. We propose VertexSerum, with data poisoning and self-attention link detector, a link inference  attack with  significantly better attack performance on intra-class nodes. We conduct extensive evaluations on different attack settings, including gray-box, offline training, online training, and black-box. 
As graph neural networks become increasingly popular, our findings pose a new challenge to confidentiality of the structural datasets using GNNs. The work serves as a cautionary note to model vendors, informing them of possible privacy exposure of their training datasets and calling for more  follow-on work to build robust GNNs against such privacy-breaching attacks.